\documentclass[manuscript,1p,times]{elsarticle}
\journal{Neural Networks}

\usepackage{times}
\usepackage{helvet}  
\usepackage{courier}  
\usepackage{url}  
\usepackage{amsmath}
\usepackage{amsfonts}
\usepackage{amssymb}
\usepackage{dsfont}
\usepackage{algorithmic}
\usepackage{algorithm}
\usepackage{hyperref}
\usepackage{graphicx}
\usepackage{epstopdf}
\usepackage{epsfig}
\usepackage{subfigure}
\usepackage{dsfont}
\usepackage{lineno}
\usepackage{multirow}
\usepackage{booktabs}
\usepackage{color}
\usepackage[flushleft]{threeparttable}
\usepackage[oldenum]{paralist}
\usepackage[deletemarkup=xout]{changes}

\usepackage{acronym}
\acrodef{OR}{Ordinal Regression}
\acrodef{ISBOR}{Incremental Sparse Bayesian Ordinal Regression}
\acrodef{SVM}{Support Vector Machine}
\acrodef{GP}{Gaussian Processes}
\acrodef{SBL}{Sparse Bayesian Learning}
\acrodef{SVOR}{Support Vector Ordinal Regression}
\acrodef{ISVOR}{Incremental Support Vector Machine for Ordinal Regression}
\acrodef{GPOR}{Gaussian Process Ordinal Regression}
\acrodef{SBOR}{Sparse Bayesian Ordinal Regression}
\acrodef{RVM}{Relevance Vector Machine}
\acrodef{RBF}{Gaussian Radial Basis Function}
\acrodef{MAP}{Maximum a Posteriori}
\acrodef{MAE}{Mean Absolute Error}
\acrodef{KDOR}{Kernel Discriminate for Ordinal Regression}
 
\newcommand{\rom}[1]{\uppercase\expandafter{\romannumeral #1\relax}}

\graphicspath{{./figures/}}

\begin{document}

\begin{frontmatter}
	
	\title{Incremental Sparse Bayesian Ordinal Regression}
	
	\author[uva]{Chang~Li\corref{mycorrespondingauthor}}
	\cortext[mycorrespondingauthor]{Corresponding author.}
	\ead{c.li@uva.nl}

	\author[uva]{Maarten de Rijke}
	\ead{derijke@uva.nl}

	\address[uva]{University of Amsterdam, Science Park 904, 1098 XH Amsterdam, The Netherlands}

\begin{abstract}
	\ac{OR} aims to model the ordering information between different data categories, which is a crucial topic in multi-label learning. 
	An important class of approaches to \ac{OR} models the problem as a linear combination of basis functions that map features to a high-dimensional non-linear space. 
	However, most of the basis function-based algorithms are time consuming. 
	We propose an incremental sparse Bayesian approach to \ac{OR} tasks and introduce an algorithm to sequentially learn the relevant basis functions in the ordinal scenario. Our method, called \ac{ISBOR}, automatically optimizes the hyper-parameters via the \emph{type-\rom{2} maximum likelihood} method. 
	By exploiting fast marginal likelihood optimization, \ac{ISBOR} can avoid big matrix inverses, which is the main bottleneck in applying basis function-based algorithms to \ac{OR} tasks on large-scale datasets. 
	We show that \ac{ISBOR} can make accurate predictions with parsimonious basis functions while offering automatic estimates of the prediction uncertainty. 
	Extensive experiments on synthetic and real word datasets demonstrate the efficiency and effectiveness of \ac{ISBOR} compared to other basis function-based \ac{OR} approaches.
\end{abstract}


\end{frontmatter}


\section{Introduction}
\label{sec:introduction}

The task of modeling ordinal data has attracted attention in various areas, including computer vision~\cite{ORCNN,ORML}, information retrieval~\cite{LTR}, recommender systems~\cite{Hu2018cfor} and machine learning~\cite{ORsurvey,elsevier1,elsevier2,elsevier3,tang2017ordinal}.
Because of the explicit or implicit relationship between labels, simple regression or multi-classification algorithms may fail to find optimal decision boundaries, which motivates the development of dedicated methods.

Generally, \ac{OR} algorithms can be classified into three categories: naive approaches, ordinal binary decompositions, and threshold models~\cite{ORsurvey}.  
For naive approaches, \ac{OR} tasks are simplified into traditional multi-classification or regression tasks, omitting ordering information, and solved by simple machine learning algorithms, e.g., \ac{SVM} Regression \cite{SVR}. 
For ordinal binary decomposition, the ordinal labels are decomposed into several binary pairs, which are then modeled by a single or multiple classifiers.  
For the threshold models, the \ac{OR} problem is addressed by training a threshold model, which models the hidden score function and an implicit set of thresholds that derive the ordinal paradigm. 
Among these three categories, the third, threshold models, is the most popular way to model the \ac{OR} problems~\cite{ORsurvey}. 
Thus, in this paper, we focus on threshold models.

Since data may lie in a low-dimensional space where data are not distinguishable by a linear combination of the features, basis functions are widely used in all three types of \ac{OR} algorithm. 
The basis function can map features to highly non-linear spaces where the data can be distinguishable by a linear combination of basis functions~\cite{vapnik1999overview}. 
We call this kind of algorithms \emph{basis function-based algorithms}. 
Most of the current basis function-based \ac{OR} algorithms do not scale well, as they are batch methods and require access to the full training dataset. 

To address this scalability problem, we propose \acfi{ISBOR}, which utilizes an incremental Bayesian approach to learning. 
We impose a zero-mean Gaussian prior over function parameters and utilize the ordinal likelihood~\cite{GPOR}, which is regarded as a probit function of \ac{OR} to model the ordinal relationship between categories. 
Then we apply the Laplace method~\cite{laplace} to derive a \ac{MAP} estimate of the unknown parameters over the dataset. 
In order to derive a full Bayesian solution, we derive a  type-II maximum likelihood optimization~\cite{RVM}, in which \ac{ISBOR} automatically optimizes the thresholds that determine the decision boundaries of ordering categories as hyper-parameters. 
Finally, to accelerate training, we follow the idea of fast marginal likelihood learning~\cite{IRVM} and derive an incremental training strategy for \ac{ISBOR}.

With this paper, we make an important step towards efficient ordinal regression based on basis functions. 
In particular, the main contributions are as follows:
\begin{itemize}
	\item We propose a basis function-based sequential sparse Bayesian treatment for ordinal regression, \ac{ISBOR}, which scales well with the number of training samples.  
	
	\item We provide an experimental evaluation of ISBOR's performance against existing basis function-based \ac{OR} algorithms in terms of efficacy, efficiency and sparseness.
\end{itemize}

\noindent%
The remainder of the paper is organized as follows. 
Section~\ref{sec:relateWork} revisits the related work. 
Section~\ref{sec:SBI} presents \ac{ISBOR}. 
Section~\ref{sec:hyper-parameter} details the hyper-parameter optimization of \ac{ISBOR}. 
We report on the experimental results in Section~\ref{sec:experimet}.
The paper is concluded in Section~\ref{sec:conclusion}. 


\section{Related Work}
\label{sec:relateWork}

In this paper, we focus on so-called basis function-based approaches to ordinal regression, which bring non-linear patterns to the linear decision functions and are well studied in machine learning.
Three types of basis function-based approaches are widely used for the \ac{OR} task: \acp{SVM}~\cite{vapnik1999overview}, \ac{GP}~\cite{gpml} and \ac{SBL}~\cite{RVM}. 
\ac{SVM} approaches convert the learning process to a convex optimization problem for which there are efficient algorithms, e.g., SMO~\cite{SMO}, to find global minima. 
However, \ac{SVM} is not equipped with a probabilistic interpretation, as a result of which it is hard to use expert or prior knowledge and make the probabilistic predictions with \ac{SVM}. 
\ac{GP}~\cite{gpml} and \ac{SBL} are Bayesian methods, which take expert knowledge as prior information and interpret the prediction with the posteriori distribution. 
In order to conduct Bayesian inference and model selection, most of them require one to compute the inverse of the basis function matrix, which leads to $\mathcal{O}(N^3)$ computational complexity, where $N$ is the number of training samples. 

In the following, we describe some of these algorithms to provide context for our work. 
The \ac{SVM}-based \ac{SVOR} approach~\cite{SVOR} is an accurate \ac{OR} algorithm~\cite{ORsurvey}. 
\ac{SVOR} is optimized using a sequential minimal optimization strategy, which brings the upper bound down to $\mathcal{O}(N^2\log N)$. 
Solving \ac{SVOR} in the dual problem boils down to optimizing with L2-regularization, which leads to a slightly sparse solution. 

\ac{ISVOR}~\cite{ISVOR} addresses the problem of basis function-based batch algorithms for \ac{OR}. 
It decomposes the \ac{OR} problem into ordinal binary classification and simultaneously builds decision boundaries with linear computational complexity. 
However, \ac{ISVOR} suffers from the problem of stability and it doubles the problem size because of its binary decomposition approach.
The main difference between the proposed \ac{ISBOR} and \ac{SVM}-based methods is that \ac{ISBOR} can use prior knowledge and make probabilistic predictions. 

\ac{GPOR}~\cite{GPOR} is the first \ac{GP} algorithm that has been proposed for the \ac{OR} task. 
\ac{GPOR} employs a \ac{GP} prior on the latent functions, and uses an ordinal likelihood, which is a generalization of the \emph{probit} function, to estimate the distribution of ordinal data conditional on the model. 
To conduct model adaptation, \ac{GPOR} applies two Bayesian inference techniques: Laplace approximation~\cite{laplace} and expectation propagation approximation~\cite{ep}, respectively. 
Since approximate Bayesian inference methods requires one to compute the inverse of an $N \times N$ matrix, the computational complexity of \ac{GPOR} is $\mathcal{O}(N^3)$. 
The main differences between \ac{GPOR} and \ac{ISBOR} are twofold: 
\begin{enumerate}
	\item \ac{ISBOR} is a sparse method, as a result of which the prediction is only based on the relevant samples. In contrast, GPOR makes predictions based on the whole training data. 
	\item \ac{ISBOR} is an incremental learning algorithm, while \ac{GPOR} is a batch algorithm: during training, \ac{GPOR} needs to compute the matrix inverse of size $N\times N$ , while \ac{ISBOR} only computes the matrix inverse of size $M \times M$, where $M \ll N$ is the number of relevant samples. 
\end{enumerate}
Based on \ac{GPOR}, various \ac{OR} algorithms have been proposed~\cite{srijith2012probabilistic,srijith2012validation,srijith2013semi,chu2005preference}. 
However, they are all batch algorithms. 
In contrast, the proposed method, \ac{ISBOR} is an incremental learning algorithm and gets rid of computing the inverse of $N\times N$ matrix. 

Based on \ac{SBL}, \ac{SBOR}~\cite{SBOR} builds a probabilistic solution to the \ac{OR} problem. 
Here, ``sparse" that means \ac{SBOR} utilizes a sparseness assumption that enables it to make predictions based on a few relevant samples with a $\mathcal{O}(M^3)$ computational bound, where $M$ is the number of relevant samples. 
However, \ac{SBOR} is still a batch algorithm and requires one to handle matrix inversion on the full dataset during initial iterations. Other basis function-based batch \ac{OR} algorithms include \ac{KDOR}~\cite{kdlor}. 

In summary, \ac{ISBOR} differs from the above algorithms in the following ways.
Instead of operating in batch, \ac{ISBOR} utilizes an incremental way to sequentially choose relevant samples. 
Because of the sparsity assumption, during sequential training \ac{ISBOR} only selects a small portion of the training data with linear computational complexity in each iteration. 
Moreover, instead of designing ordinal partitions like \ac{ISVOR}, \ac{ISBOR} directly learns the implicit thresholds and score function, which is a more natural way to reveal ordinal relations.


\section{\acl{ISBOR}}
\label{sec:SBI}
We start this section by defining the notation used in the paper. 
The training set is $\mathcal{D} = \{\mathbf{x}_n,y_n\}_{n=1}^N$, where $\mathbf{x}_n \in \mathbb{R}^d$ is the feature vector, $y_n \in \{1,2,\ldots,r\}$ is the corresponding category; $r$ is the number of categories. 
We use normal-face letters to denote scalar and boldface letters to denote vectors and matrices.

We present \ac{ISBOR} in four steps: model specification, likelihood definition, prior assumption and maximum a posterior. 

\subsection{Model specification}
\label{sec:Model}
As a threshold \ac{OR} model \cite{ORsurvey}, \ac{ISBOR} chooses a linear combination of basis functions as the score function, $f(\mathbf{x}_n;\mathbf{w})$, which maps a sample from the $d$-dimensional feature space to a real number: 
\begin{equation}\label{eq:model}
f(\mathbf {x}_n) = \sum_{i=1}^N \phi_i(\mathbf{x}_n) w_i = \boldsymbol \phi(\mathbf{x}_n) \mathbf{w}, 
\end{equation}
where $\mathbf{w} \in \mathcal{R}^N$ denotes the parameter vector\footnote{Here, $w_n$ controls the relevance of the $n$-th basis function $\phi_n(\mathbf{w})$: if $w_n =0$, the $n$-th basis function is irrelevant for the decision, which is equivalent to throw the $n$-th sample away and retain the relevant basis functions.} and $\boldsymbol \phi(\mathbf{x}_n) = [\phi_1(\mathbf{x_n}),\ldots,\phi_N(\mathbf{x_n})]$ is the basis function, e.g., the \ac{RBF}: 
\begin{equation} \label{eq: kernel}
\phi(\mathbf{x_n},\mathbf{x_i}) = \exp{(-\theta{\|\mathbf{x}_n-\mathbf{x}_i\|_2^2})}. 
\end{equation}
After mapping, \ac{ISBOR} exploits a set of thresholds, $[b_0,\ldots,b_r]$, to determine intervals of different categories. In order to represent the ordering information, these thresholds are chosen as a set of ascending numbers, e.g., $b_{i+1} > b_i$, and work with a set of positive auxiliary numbers, $[ \Delta_2,\ldots,\Delta_{r-1}]$, with $b_n$ defined as $b_n = b_1 + \sum_{i=2}^n \Delta_i$. 
During prediction, a sample $\mathbf{x}_n$ is classified to a target $y_n$ if and only if $b_{y_n-1} < f(\mathbf{x}_n) \leq b_{y_n}$. 
We set $b_0 = -\infty$ and $b_r = \infty$.

\subsection{Ordinal likelihood}
To model ordinal data, we take the ordinal likelihood proposed in \ac{GPOR}~\cite{GPOR}. 
The likelihood is the joint distribution of the samples conditional on the model parameters, and with the I.I.D.\ assumption; it is computed as: 
\begin{equation*}
p(\mathbf{Y}\mid \mathbf{X},\mathbf{w}) = \prod_{n=1}^N p(y_n\mid \mathbf{X},\mathbf w),
\end{equation*}
where $\mathbf{Y} = \{y_n\}_{n=1}^N$ and $\mathbf{X} = \{\mathbf {x}_n\}_{n=1}^N$. 
Following the standard probabilistic assumption~\cite{GPOR}, we assume that the outputs of a score function are contaminated with random Gaussian noise: $\hat{y}_n = f(\mathbf{x}_n)+\epsilon$, where $\epsilon \sim \mathcal{N}(0,\sigma^2)$. 
$\sigma$ is the standard deviation of the noise distribution, which is learned by the model selection (Section~\ref{sec:threshold}). 
In this way, the score function is linked to the probabilistic output $p(\hat{y}_n\mid \mathbf{w},\mathbf {x}_n, \epsilon) = \mathcal{N}(\hat{y}_n\mid f(\mathbf{x}_n),\sigma^2)$. 
And the likelihood over a sample is  computed as follows: 
\begin{equation}\label{eq:idea}
p_{ideal}( y_n\mid \mathbf{x}_n,\mathbf w, \epsilon) = 
\begin{cases}{}
1 \qquad \text{if }b_{y_n-1} < \hat{y}_n \leq b_{y_n},\\
0 \qquad \text{otherwise}.
\end{cases}
\end{equation}
Since $b_{i+1} = b_i + \delta_{i+1}$ and $\delta_{i+1} > 0$, $[b_0,\ldots,b_r]$ divide the real line into $r$ ordinal intervals. 
Thus, with these intervals, the ideal likelihood maps the real value output $f(x)$ to ordinal categories. 
However, because of the uniform distribution, Eq.~\eqref{eq:idea} is not differentiable, and hence we cannot implement Bayesian inference. 
To tackle this issue, we integrate out the noise term and obtain a differentiable likelihood as follows: 
\begin{equation} 
\begin{split}
p(y_n\mid{} \mathbf{x}_n,\mathbf {w}, \sigma) 
&= \int_{\epsilon} \!p_{ideal}({y}_n\mid \mathbf{x}_n,\mathbf w, \epsilon) \mathcal{ N}(\epsilon\mid 0,\sigma^2) d \epsilon \\
& = \psi(z_{n,1}) - \psi(z_{n,2}),
\end{split}
\label{eq:likelihood} 
\end{equation}
where 
\[
z_{n,1} = \frac{b_{y_n} - f(\boldsymbol x_n)}{\sigma}
\text{ and }
z_{n,2} = \frac{b_{y_n-1} - f(\mathbf x_n)}{\sigma},
\]
and $\psi(z)$ is the Gaussian cumulative distribution function. 
Based on Eq.~\eqref{eq:likelihood}, maximum likelihood estimation is equivalent to maximizing the area under the standard Gaussian distribution between $z_{n,1}$ and $z_{n,2}$, which is differentiable. 

\subsection{Priori assumption}
For large scale datasets, if we directly learn parameters by maximum likelihood estimation, we may easily encounter severe over-fitting. 
To avoid this, we add an additional constraint on parameters: the regularization term. 
In Bayesian learning, we achieve this by introducing a zero-mean Gaussian prior for $\mathbf{w}$: $p(w_n\mid \alpha_n) = \mathcal{N}(w_n;0,\alpha_n^{-1})$. 
Assuming that each parameter is mutually independent, the prior over parameters is computed as: 
\begin{equation} \label{eq:prior}
p(\mathbf w\mid \boldsymbol{\alpha}) = \prod_{n=1}^N \mathcal{N}(w_n\mid 0,\alpha_n^{-1}),
\end{equation}
where $\boldsymbol{\alpha} = [\alpha_1,\ldots,\alpha_N]$ and $\alpha_n$, the inverse of variance, serves as the regularization term.
If the value of $\alpha_n$ is large, the posterior of $w_n$ will be mainly constrained by the prior and $w_n$ will be bound to a small neighborhood of $0$.\footnote{Practically, when $w_n$ is smaller than a value, e.g., $10^{-3}$, we will consider it to be $0$, which boils down to throwing away the corresponding sample.} 
To complete the definition of the sparse prior, we define a set of flat Gamma hyper-priors over $\boldsymbol{\alpha}$, which together with Gaussian priors result in Student's-t prior and work as $L_1$ regularization~\cite{RVM}.

\subsection{Maximum a posterior}
\label{sec:SBL:MAP}

\begin{figure*}[t]
	\centering
	\subfigure{
		\includegraphics[width = 0.32\textwidth]{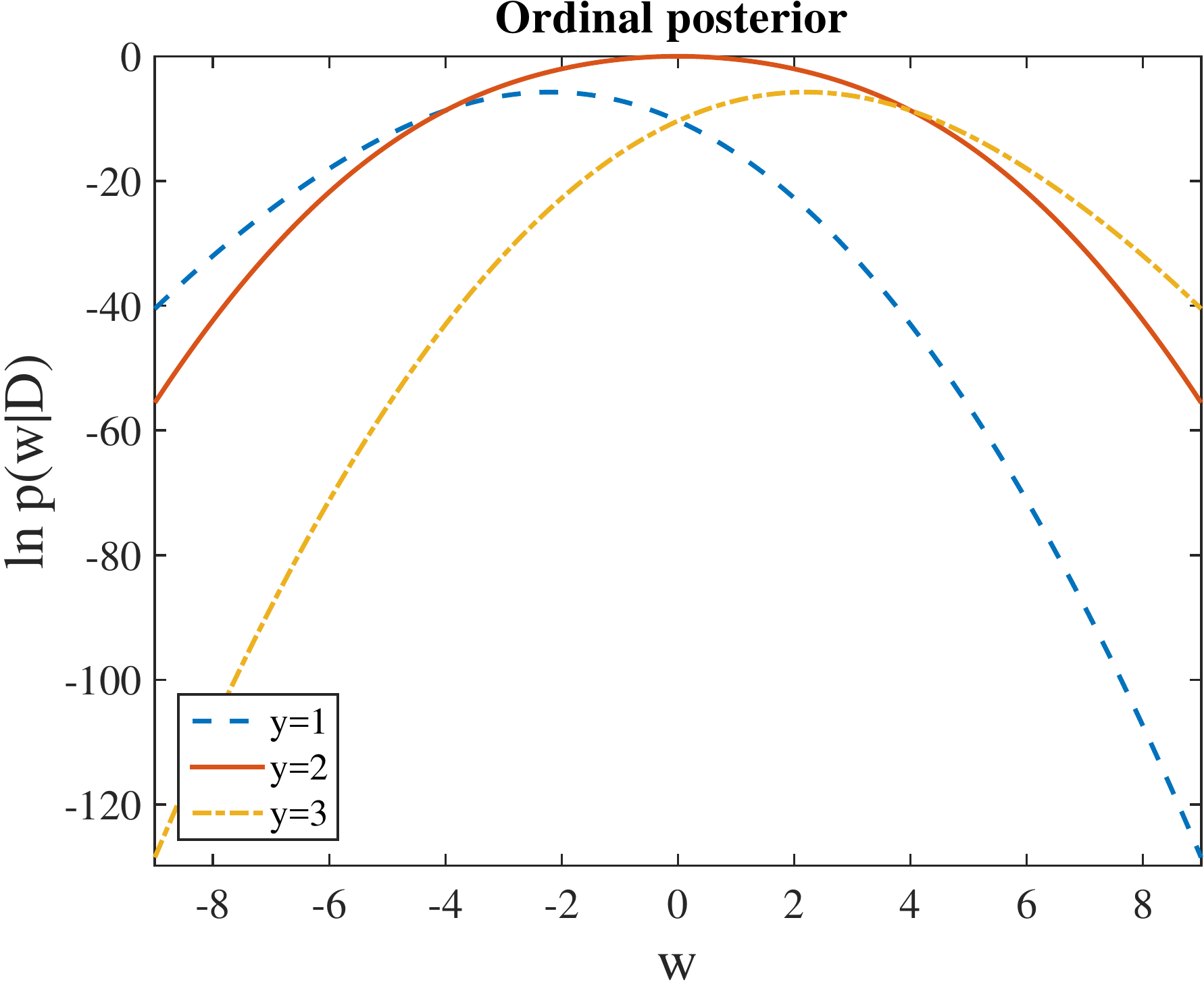}}
	\subfigure{
		\includegraphics[width = 0.32\textwidth]{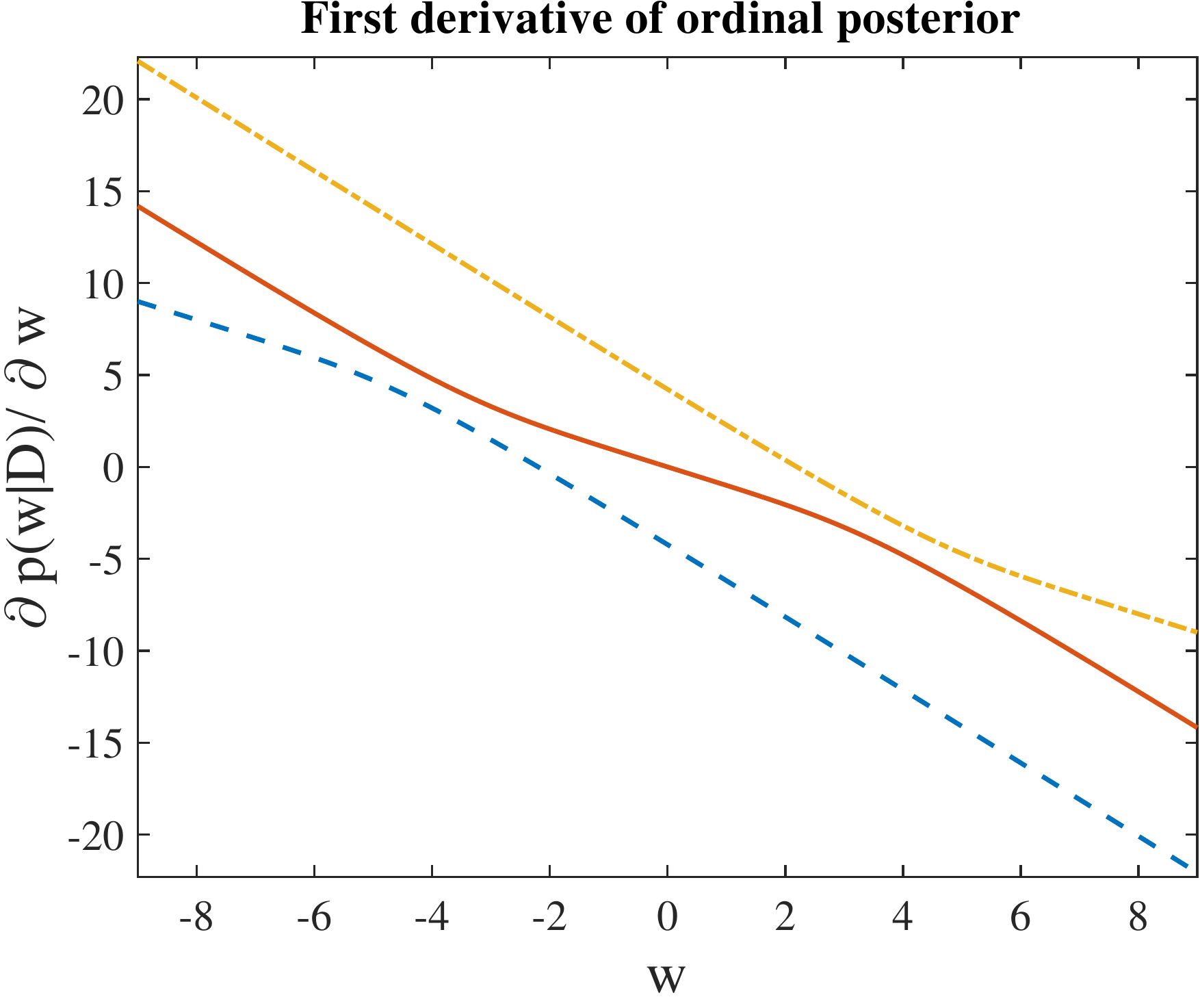}}
	\subfigure{
		\includegraphics[width= 0.32\textwidth]{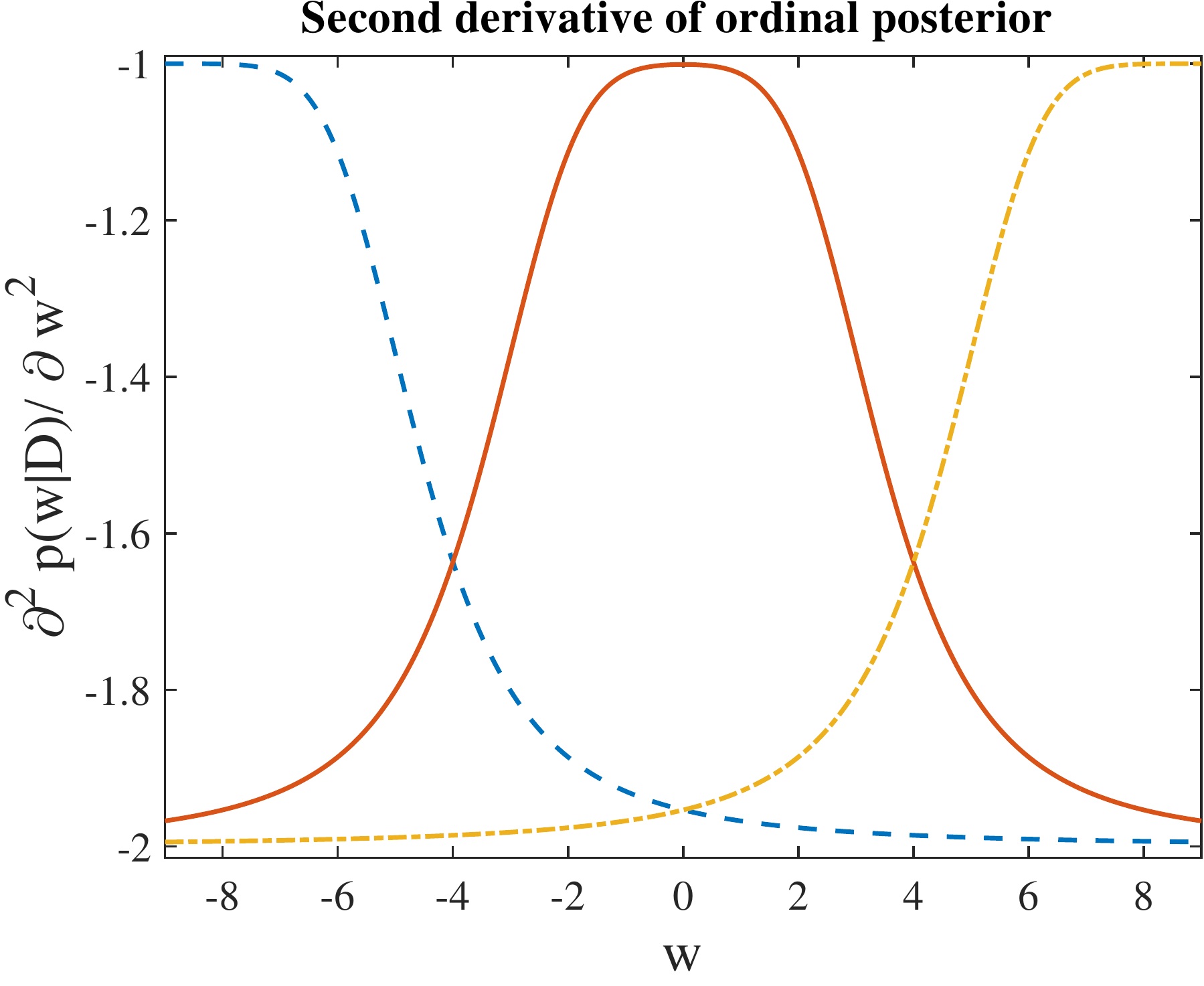}}
	
	\centering
	\caption{The ordinal posterior and its first and second derivatives.}
	\label{fig:post}
\end{figure*}

Having defined the prior and likelihood, \ac{ISBOR} proceeds by computing the posterior over all training data, based on Bayes' rule: 
\begin{equation}
\label{eq:posterior}
p(\mathbf w\mid \mathbf D) = \frac{p(\mathbf{Y}\mid \mathbf{X},\mathbf{w},\sigma) p(\mathbf w\mid \boldsymbol{\alpha})}{p(\mathbf D\mid \boldsymbol{\eta})} ,
\end{equation}
where $\mathbf{D}$ is the training data set, $p(\mathbf{w}\mid \mathbf{\alpha})$ defined in Eq.~\eqref{eq:prior} is the prior, $p(\mathbf{Y}\mid \mathbf{X},\mathbf{w},\sigma)$ defined in Eq.~\eqref{eq:likelihood} is the likelihood, the denominator $p(\mathbf D\mid \boldsymbol{\eta}) = \int p(\mathbf{Y}\mid \mathbf{X},\mathbf{w},\sigma) p(\mathbf{w}\mid \boldsymbol{\alpha})d\mathbf{w} $ is the marginal likelihood, which we use for model selection and hyper-parameter optimization in the next section. 
To simplify our notation, we collect all the hyper-parameters, including noise level $\sigma$, thresholds and $\boldsymbol{\alpha}$, into $\boldsymbol{\eta}$. 

We prefer the $\mathbf{w}^*$ with the highest posterior probability, and formulate the \ac{MAP} point estimate as $\mathbf{w}^* = \max_{\mathbf {w}} p(\mathbf w\mid \mathbf D)$. 
However, we cannot integrate $\mathbf{w}$ out in the marginal likelihood analytically. 
In our \ac{MAP} estimation we use the fact that $p(\mathbf w\mid \mathbf D) \propto p(\mathbf{Y}\mid \mathbf{X},\mathbf{w},\sigma) p(\mathbf w\mid \boldsymbol{\alpha})$ and work with the logarithm of the posterior:
\begin{equation}
\begin{split}
\ln p(\mathbf{w}\mid {} \mathbf{D})
 &= \ln p(\mathbf{Y}\mid \mathbf{X},\boldsymbol w,\sigma)+ \ln p(\mathbf w\mid \boldsymbol \alpha)
+const \\
&\approx \sum_{n=1}^N\ln(\psi(z_{n,1})-\psi(z_{n,2})) - \frac{1}{2} \mathbf{w}^T\mathbf{A} \mathbf{w},
\end{split}
\label{eq:map}
\end{equation}
where $\mathbf A$ is a diagonal matrix with diagonal elements $[\alpha_1,\ldots,\alpha_N]$, $const$ is a term independent of $\mathbf w$. 
The first part of the last line, from the likelihood, works as the loss term; the second part, from the prior, acts as the regularization term. 

Next, the Newton-Raphson method \cite{newton} is applied to compute the \ac{MAP} estimate.
First, we compute the first and second order derivatives of the first term (log-likelihood part), $ \mathcal{L}= \ln p(\mathbf{Y}\mid \mathbf{X},\mathbf w)$: 
\begin{eqnarray}
\frac{\partial \mathcal{L}}{\partial \mathbf w} &=& -\sum_{n=1}^N\frac{1}{\sigma}\frac{\mathcal{ N}(z_{n,1}\mid 0,1) - \mathcal{N}(z_{n,2}\mid 0,1)}{\psi(z_{n,1})-\psi(z_{n,2})}\boldsymbol\phi_n \notag\\
& =& \boldsymbol{\Phi}^T \delta \label{eq:dl}\\
\frac{\partial^2 \mathcal{L}}{\partial \mathbf{w} \partial \mathbf{w}^T} &=& -\boldsymbol\Phi^T \mathbf H \boldsymbol\Phi, \label{eq:ddl}
\end{eqnarray}
where 
\begin{eqnarray*}
		\delta_n &=&\frac{1}{\sigma}\frac{\mathcal{ N}(z_{n,1}\mid 0,1) - \mathcal{N}(z_{n,2}\mid 0,1)}{\psi(z_{n,1})-\psi(z_{n,2})} \\
	H_{nn} &=& \frac{1}{\sigma^2}\left[\left(\frac{N(z_{n,1}\mid 0,1) - N(z_{n,2}\mid 0,1)}{\psi(z_{n,1})-\psi(z_{n,2})}\right)^2 \frac{z_{n,1}N(z_{n,1}\mid 0,1) - z_{n,2}N(z_{n,2}\mid 0,1)}{\psi(z_{n,1})-\psi(z_{n,2})}\right].
\end{eqnarray*}
Then, combining Eq.~\eqref{eq:map}, Eq.~\eqref{eq:dl} and Eq.~\eqref{eq:ddl}, we obtain the derivative of the log-posterior as \[
\frac{\partial^2 \log{p(\mathbf{w}\mid \mathbf{D})}}{\partial \mathbf{w}\partial \mathbf{w}^T} = -\boldsymbol\Phi^T \mathbf H \boldsymbol\Phi-\mathbf{A}.
\]
Note that 
$\boldsymbol\Phi^T \mathbf {H} \boldsymbol\Phi$ is a quadratic form and $\mathbf{A}$ is a diagonal matrix with positive diagonal elements, so 
\[
- \frac{\partial^2 \log{p(\mathbf{w}\mid \mathbf{D})}}{\partial \mathbf{w}\partial \mathbf{w}^T}
\]
is a positive definite matrix, which implies that MAP estimation is a concave programming problem, with a global maximum.

Having found the \ac{MAP} point $\mathbf{w}^*$, we use the Laplace method to approximate the posterior distribution by a Gaussian distribution $\mathcal{N}(\mathbf{w}\mid \mathbf{w}^*,\boldsymbol{\Sigma})$, where $\mathbf{w}^*$ and $\boldsymbol{\Sigma}$ are the mean and variance and computed as follows:
\begin{eqnarray}
\boldsymbol \Sigma &=& (\mathbf A+ \boldsymbol\Phi^T \mathbf H \boldsymbol\Phi)^{-1} \label{eq:variance}\\
 \mathbf{w}^* &=& \boldsymbol{\Sigma}\boldsymbol{\Phi}^T\mathbf{H}\hat{\mathbf{t}}, \label{eq:MAP}
\end{eqnarray}
where $\mathbf{\hat{t}} = \mathbf{H}^{-1} \boldsymbol{\delta} + \boldsymbol{\Phi}\mathbf{w}^*$. 

Using a local Gaussian at the \ac{MAP} point to represent the posterior distribution over weights is often considered as a weakness of the Bayesian treatment, especially for complex models. 
However, as pointed out by Tipping~\cite{RVM}, a log-concave posterior implies a much better accuracy and no heavier sparsity than L1-regularization. 
As we discussed above, the posterior of \ac{ISBOR} has the feature of log-concavity. 
We report the plots of the log-posterior as well as its first and second order derivatives in Figure~\ref{fig:post} and see that $\frac{\partial \mathcal{L}}{\partial \mathbf w} $ is monotonically decreasing w.r.t.\ $w$, while $\frac{\partial^2 \mathcal{L}}{\partial^2 \mathbf w}$ is always smaller than $0$. 
So the \ac{MAP} here is essentially a log-concave optimization problem, which implies that the Laplace approximation in \ac{ISBOR} enjoys the same features of accuracy and sparsity as in the \ac{RVM}~\cite{RVM}. 
	

\section{Hyper-parameter Optimization}
\label{sec:hyper-parameter}

\ac{ISBOR} uses various hyper-parameters, including $\boldsymbol{\alpha}$ in the prior estimation (Eq.~\eqref{eq:prior}), the noise variance $\sigma$ in Eq.~\eqref{eq:likelihood}, and the thresholds $[b_1,\ldots,b_r]$. 
In this section we detail how to learn these hyper-parameters. 

\subsection{Marginal likelihood}

As a fully Bayesian framework,  hyper-parameters are optimized by maximizing the posterior mode of hyper-parameters $p(\boldsymbol{\eta}\mid \mathbf{D}) \propto p(\mathbf{D}\mid  \boldsymbol{\eta} ) p(\boldsymbol{\eta})$, where $\boldsymbol{\eta}$ contains all hyper-parameters. 
As we  assume a non-informative Gamma hyper-prior, the optimization is equivalent to maximizing the marginal likelihood  $p(\mathbf{D}\mid  \boldsymbol{\eta} )$, which is computed as $p(\mathbf{D}\mid \boldsymbol{\eta}) = \int p(\mathbf{D}\mid \mathbf{w},\sigma) p(\mathbf{w}\mid \boldsymbol{\alpha})d \mathbf{w}$. 
As there is no closed form for this equation, again, we apply Laplace approximation and get the following approximations: 
\begin{equation}
\begin{split}
p(\mathbf D\mid \boldsymbol \eta) &= p(\mathbf{Y}\mid \mathbf{w}^*) p(\mathbf{w}^*\mid \boldsymbol \alpha) (2\pi)^{n/2}\boldsymbol{\Sigma}^{1/2} \\
\ln p(\mathbf D\mid \boldsymbol \eta) &= \mathcal{L} - \frac{1}{2} \mathbf w^{*T}\mathbf{A}\mathbf w^* +\frac{1}{2}\ln|\mathbf{A}| + \frac{1}{2}\ln|\boldsymbol \Sigma|. 
\end{split}
\label{eq:ml}
\end{equation}
In the rest of this section, we deal with the log-marginal likelihood, and maximize  Eq.~\eqref{eq:ml} with respect to each hyper-parameter. 

\subsection{Threshold and noise hyper-parameters}
\label{sec:threshold}
For the threshold hyper-parameters, we only need to determine $r-1$ values: $b_1$ and $[\Delta_2$, \ldots, $\Delta_{r-1}]$.  
Since we cannot compute these  analytically, we exploit gradient descent (ascent, actually) to iteratively choose these parameters. 
The derivatives of the log-marginal likelihood, Eq.~\eqref{eq:ml}, with respect to $b_1$ and $\Delta_{i}$, are computed as follows:
\begin{eqnarray} 
\frac{\partial \ln p(\boldsymbol D\mid \boldsymbol \eta)}{\partial b} &=& -\boldsymbol \delta^*, \label{eq:b}\\
\frac{\partial\ln p(\boldsymbol D\mid \boldsymbol \eta)}{\partial \Delta_i} &=& \begin{cases}
-\delta_n^* & \text{ if }y_n > i \\
\frac{1}{\sigma}\frac{ \mathcal{N}(z_1;0,1)}{\Psi(z_1)-\Psi(z_2)} & \text{ if } y_n = i\\
0 &~ \text{otherwise}.
\end{cases} \label{eq:Delta}
\end{eqnarray}
Based on these two equations, we use gradient descent to search for proper thresholds. 

For the noise term $\sigma$, setting the derivative 
\[
\frac{\ln p(\mathbf{D}\mid \boldsymbol \eta)}{\sigma} = 0,
\]
we obtain an update rule for the noise term: 
\begin{equation}
\sigma^2 = \frac{\|\mathbf{\hat{t}}- \boldsymbol{\Phi w}\|^2}{N - \sum_{n}(1-\alpha_n \Sigma_{nn})},
\label{eq:noise}
\end{equation}
where $\mathbf{\hat{t}} = \mathbf{H}^{-1} \boldsymbol{\delta} + \boldsymbol{\Phi}\mathbf{w}^*$. 

\subsection{Fast marginal learning}
We compute the contribution of the sparsity hyper-parameter $\boldsymbol{\alpha}$ to the marginal likelihood as follows: 
\begin{equation}\label{eq:mlalpha}
\ln p(\mathbf D\mid \boldsymbol \alpha) = \mathcal{L}-\frac{1}{2}\ln|\mathbf{C}| - \frac{1}{2}\hat{t}\mathbf{C}^{-1}\hat{t},
\end{equation}
where we compute $\mathbf{C}$ as follows: 
\begin{equation}
\begin{split}
\mathbf{C} &= \mathbf{H}^{-1} + \boldsymbol{\Phi} \mathbf{A}^{-1}\Phi^T\\
&= \mathbf{H}^{-1} + \sum_{n\neq j}\alpha_n\boldsymbol \phi_n \boldsymbol \phi_n^T + \alpha_j^{-1}\boldsymbol\phi_j\boldsymbol\phi_j^T.
\end{split}
\label{eq:C}
\end{equation}
Since computing $\mathbf{C}$ requires matrix inversion, it is impractical to maximize it for large scale training sets. 
Fortunately, Tipping and Faul~\cite{IRVM} proposed a sequential way to maximize the marginal likelihood. 
We take this strategy and optimize $\boldsymbol{\alpha}$ as follows:

\begin{itemize}
	\item First, we use the established matrix determinant and inverse identities~\cite{matrix_cookbook} to compute the determination and inverse of $\mathbf{C}$ as follows:
	\begin{equation}
	\begin{split}
	|\mathbf{C}| &= |\mathbf{C}_{/j}|| \mathbf{I} + \alpha_j^{-1}\boldsymbol{\phi}_j\boldsymbol{\phi}_j^T| \\
	\mathbf{C}^{-1} &= \mathbf{C}_{/j}^{-1} - \frac{\mathbf{C}_{/j}^{-1}\boldsymbol{\phi}_j\boldsymbol{\phi}_j^T\mathbf{C}_{/j}^{-1}}{\alpha_j + \boldsymbol{\phi}_j^T\mathbf{C}_{/j}^{-1}\boldsymbol{\phi}_j}, 
	\end{split}
	\label{eq:c_decompose}
	\end{equation} 
	where $\mathbf{I}$ is the identity matrix, and $\mathbf{C}_{/j}$ denotes $\mathbf{C}$ without the contribution of the $j$-th sample. 
	
	\item Second, we define two auxiliary variables: 
	\begin{equation}\label{eq:qs}
	s_j = \phi_j^TC_{/j}^{-1}\phi_j, ~q_j = \phi_j^TC_{/j}^{-1}\hat{t}.
	\end{equation}
	Combining Eqn.~\eqref{eq:mlalpha}, \eqref{eq:c_decompose} and~\eqref{eq:qs}, we isolate the contribution of sample $j$ to the marginal likelihood as follows: 		
	\begin{equation}\label{eq:singleML}
	\ln p(\boldsymbol D\mid \alpha_j) = \frac{1}{2}[\ln \alpha_j - \ln |\alpha_j + s_j| + \frac{q_j^2}{s_j+\alpha_j}]. 
	\end{equation}
	For simplicity, we define $g(\alpha_j) = \ln p (\boldsymbol D\mid \alpha_j)$.
	
	\item However, we still need to compute the inverse of $\mathbf{C}_{/j}$ in Eq.~\eqref{eq:qs}. To speed up the computation, we define the follow auxiliary variables: 
	\begin{eqnarray*}
		\begin{split}
			Q_j ={}&\boldsymbol{ \phi}_j^T \mathbf{C}^{-1}\hat{t}  =\boldsymbol{ \phi}_j^T \mathbf{H} \hat{t} - \boldsymbol{\phi}_j^T \mathbf{H} \boldsymbol{\Phi} \boldsymbol{\Sigma} \boldsymbol{\Phi}^T \mathbf{H} \hat{t} \\
			S_j = {}&\boldsymbol{ \phi}_j^T \mathbf{C}^{-1} \boldsymbol{\phi}_j = \boldsymbol{\phi}_j^T \mathbf{H} \boldsymbol{\phi}_j - \boldsymbol{\phi}_j^T \mathbf{H} \boldsymbol{\Phi} \boldsymbol{\Sigma} \boldsymbol{\Phi}^T \mathbf{H} \boldsymbol{\phi}_j, 
		\end{split}
		\label{eq:QS}
	\end{eqnarray*}
	where $\boldsymbol{\Sigma} \in \mathbb{R}^{M \times M}$ is the covariance of the posterior distribution (Eq~\eqref{eq:variance}).\footnote{ Because of the sparse assumption, $M \ll N$, and 
	thus computing the inverse of $\boldsymbol{\Sigma}$ is much faster than that of $\mathbf{C}$.} 
	Then, we can compute 
	\[
	s_j = \frac{\alpha_jS_j}{\alpha_j - S_j}\text{ and }q_j = \frac{\alpha_j Q_j}{\alpha_j-S_j}.
	\]

	\item Finally, setting 
	\[
	\frac{\partial g(\alpha_j)}{\partial \alpha_j}  = 0,
	\]
	we get the closed form solution for $\alpha_j$:  
	\begin{equation}
	\alpha_j = \frac{s_j^2}{q_j^2-s_j}. \label{eq:alpha}  
	\end{equation}
	Since  $\alpha_j\geq0$, the denominator of Eq.~\eqref{eq:alpha}, denoted as $f_j = q_j^2-s_j > 0$, which works as an important criterion for determining the relevant samples.	
\end{itemize}

\subsection{\ac{ISBOR}}
We summarize the pseudo-code of \ac{ISBOR} in Algorithm~\ref{alg:sbor}. We provide brief comments on three ingredients. 
First, we initialize \ac{ISBOR} (line~4) by randomly picking a sample from each category as the initial relevant samples.
Based on these $r$ samples, we initialize $\mathbf{Q}$, $\mathbf{S}$ and $\mathbf{f}$. On Line~6, we compute the delta marginal likelihood for the samples not yet considered. 
As to the call to Estimate() (line~13), we update $\mathbf{w}$ based on Eq.~\eqref{eq:MAP}; update $\boldsymbol{\alpha}$ based on Eq.~\eqref{eq:alpha}; update $\mathbf{ml}$ based on Eq.~\eqref{eq:ml} and use gradient search to update threshold $\mathbf{b}$ based on Eq.~\eqref{eq:b} and Eq.~\eqref{eq:Delta}.

\begin{algorithm}[!t]
	\caption{\acf{ISBOR}}
	\label{alg:sbor}
	\begin{algorithmic}[1]
		\STATE \textbf{Input:} $\mathbf{D} = \{\mathbf{x},\mathbf{y}\}, \theta$, maxIts and minDelta.
		\STATE \textbf{Output:} $\mathbf{w}$, $\mathbf{b}$ and $\sigma$.
		\STATE $\boldsymbol{\Phi}$ = basis($\mathbf{x}, \theta$);
		\STATE $\mathbf{w}, \boldsymbol{\phi}, \boldsymbol{\alpha}, \sigma,\mathbf{b,Q,S,f} = $ Initialize$(\Phi, \mathbf{y})$;
		\FOR{$i=1,2,\ldots,\text{maxIts}$ }
		\STATE  deltaML = $[g(\alpha_1), \ldots, g(\alpha_n)]$;
		\STATE $\boldsymbol{\phi_n}	 \leftarrow$ max(deltaML);
		\IF{$\boldsymbol{\phi_n}\in \boldsymbol{\phi}$ and $f_n < 0$}
		\STATE  $\{\mathbf{w},\boldsymbol{\alpha},\boldsymbol{\phi}\} \leftarrow \{\mathbf{w},\boldsymbol{\alpha},\boldsymbol{\phi} \}-\{w_n,\alpha_n,\phi_n\}$;
		\ELSIF{$f_n>0$}
		\STATE $\{\mathbf{w},\boldsymbol{\alpha},\boldsymbol{\phi}\} \leftarrow \{\mathbf{w},\boldsymbol{\alpha},\boldsymbol{\phi} \}\cup\{w_n,\alpha_n,\phi_n\}$;
		\ENDIF
		\STATE $\mathbf{w},\boldsymbol{\alpha},\mathbf{b},\mathbf{ml}=$ Estimate($\mathbf{w},\boldsymbol{\alpha},\mathbf{b},\boldsymbol{\Phi},\boldsymbol{\phi},\sigma$);
		\STATE compute $\sigma$ based on Eq.~\eqref{eq:noise};
		\STATE compute $\boldsymbol{Q,S}$ based on  Eq.~\eqref{eq:QS};
		\STATE compute $\boldsymbol{q,s, f}$;
		\IF{abs$(\mathbf{ml}-\mathbf{ml}_{old}) <$ minDelta}
		\STATE break;
		\ENDIF		
		\STATE $\mathbf{ml}_{old} = \mathbf{ml}$;
		\ENDFOR
	\end{algorithmic}	
\end{algorithm}

\subsection{Computational analysis}

The maximization rule for marginal likelihood is based on the \ac{MAP} estimate which, in Eq.~\eqref{eq:variance}, requires the inversion of a matrix with $\mathcal{O}(M^3)$ computational complexity and $\mathcal{O}(M^2)$ memory. 
However, as we constructively maximize the marginal likelihood, $M \ll N$, first, we choose one sample from each category to initialize the algorithm; second, we benefit from the sparse learning, as the scale of $M$ remains small (around a few dozen based on our experiments). 
In this case, matrix inversion is not the main computational bottle-neck for each iteration.

Although we apply an incremental strategy to train \ac{ISBOR}, we have to compute the basis function matrix in the initialization step, which has $\mathcal{O}(N^2)$ computational complexity and $\mathcal{O}(N^2)$ memory. Combining these two parts, the total computational complexity of \ac{ISBOR} is $\mathcal{O}(N^2+M^3)$ and the memory complexity $\mathcal{O}(N^2)$. 
However, we should mention that the basis function matrix can be computed in the pre-training session, so the computational complexity is essentially $\mathcal{O}(N + M^3)$. 
For comparison, we report the computational and space complexity of \ac{SBOR} and other state-of-the-art methods in Table~\ref{tb:complexity}. 
We see that \ac{ISBOR} has the best computational complexity, and thus, \ac{ISBOR} is more efficient than others, at least theoretically. 

\begin{table}
	\caption{Computational and space complexity of ordinal regression algorithms. $N$ and $M$ represent the number of training samples and the number of relevant and/or support samples respectively. }
	\label{tb:complexity}
	\centering
	\begin{tabular}{@{~~} l @{~} c @{~~} r @{~~} r @{~~}}
		\toprule
		 	& \ac{KDOR}/\ac{GPOR}/ & \ac{ISVOR} & \ac{ISBOR} \\ 
		 	& SVOR/\ac{SBOR} &  &  \\ 
		\midrule
		Computational complexity  	 	&$\mathcal{O}(N^3)$ & $\mathcal{O}(2N + 8M^3)$ & $\mathcal{O}(N + M^3)$ \\
		Space complexity		&$\mathcal{O}(N^2)$ & $\mathcal{O}(4N^2)$          & $\mathcal{O}(N^2)$	\\
		\bottomrule
	\end{tabular}
\end{table}

As computing the posterior covariance requires the inverse of the Hessian matrix, $(\mathbf A+  \boldsymbol\Phi^T \mathbf H \boldsymbol\Phi)^{-1}$, it is inevitable to encounter the singular values. 
Theoretically speaking, $\mathbf{H}$ and $\mathbf{A}$ are the diagonal matrices with positive elements, $ \boldsymbol\Phi^T \mathbf H \boldsymbol\Phi$ is the quadratic form. 
However, there still exist singular problems, especially when some $\alpha$ are extremely large. 
In order to avoid ill-conditioning, we manually prune training samples with large $\alpha$ values. 

\subsection{Sparsity analysis}
The simple Gaussian prior working as an L2-regularization in the posterior model leads to a non-sparse \ac{MAP} estimate. 
However, with the Gamma hyper-prior, the real prior over $\mathbf{w}$ follows a Student's t distribution which is considered as a sparse prior with a sharp peak at $0$~\cite[Section 5.1]{RVM}.  
During inference, we do not integrate out $\boldsymbol{\alpha}$, which implies that $\boldsymbol{\alpha}$ is the direct factor to sparsity, which in turn means that for irrelevant vectors the corresponding $\boldsymbol{\alpha}$ should be large. 
However, the learned $\boldsymbol{\alpha}$ in the sequential model are relatively small: we only add potentially relevant samples whose $\boldsymbol{\alpha}$ are essentially small to the model. 
There is no reason to learn $\boldsymbol{\alpha}$ of samples excluded from the model, which have large values. 


\section{Experimental Evaluation}
\label{sec:experimet}
Our experimental evaluation aims at addressing the following three research questions. 
\begin{enumerate}[\bf RQ1]
	\item Efficacy: Is the generalization performance of the proposed algorithm, \ac{ISBOR}, comparable to other baselines?
	\item Efficiency: Does fast marginal analysis reduce \ac{ISBOR}'s computational complexity compared to baselines? 
	\item Sparseness: Can \ac{ISBOR} achieve the competitive predictions only based on a small subset of the training set?
\end{enumerate}

\subsection{Experimental design}
The research questions listed above lead us to two experimental designs.
The first involves a synthetic dataset to give us an understanding of the efficacy, effectiveness and sparsity. 
The second is on benchmark datasets, i.e., $7$ widely used ordinal datasets to extensively evaluate the performance of \ac{ISBOR}. 

\subsubsection{Datasets}
\paragraph{Synthetic dataset}
To create a synthetic dataset we follow the data-generating strategy in \cite{toy}. 
First, $21,000$ two-dimensional points are sampled within the square area $[0,10] \times [0,10] $ under a uniform distribution. 
Second, each point is assigned a score by the function $f(\mathbf{x}) = 10(x_1-0.5)(x_2-0.5) + \epsilon$, where $\epsilon \sim \mathcal{N}(0,0.5^2)$ acts as a Gaussian random noise. 
Finally, we choose six thresholds $\{-\infty$, $-60$, $-9$, $15$, $60$, $+\infty\}$, and each point is attached with a category by computing:
\begin{equation*}
y = \operatorname*{arg\,min}_{r\in\{1,2,3,4,5\}} b_{r-1}\leq 10(x_1-0.5)(x_2-0.5) + \epsilon \leq b_{r}.
\end{equation*}
In this manner, we generate a five-category dataset and the numbers of data points assigned to each category are $4431$, $4535$, $3949$, $3780$ and $4305$, 
respectively.
We choose $10$ different sizes of training sets: $1000$, $2000$, \ldots, $10000$ and use the rest of the data as test sets. 
For each size training sets, we randomly generate $30$ different partitions. 
Then, the experiments are conducted on all $30$ partitions. 

\paragraph{Benchmark datasets}
We also compare \ac{ISBOR} with five algorithms on seven benchmark datasets.\footnote{\url{http://www.uco.es/grupos/ayrna/ucobigfiles/datasets-orreview.zip}}
The details of the benchmark datasets are summarized in Table~\ref{tb:benchmarks}. 
\begin{table}[h]
	\caption{Benchmarks: Detailed information.}
	\label{tb:benchmarks}
	\centering
	\begin{tabular}{l r r r r} 
		\toprule
		Dataset 	& \# Training 	& \# Test & \# Features & \# Categories \\ 
		\midrule
		BS 	 	&468		&157	&4 &3\\
		SWD		&750		&250 &10 &4	\\
		Marketing & 6,744 & 2,249 &74 &9\\
		Bank 	& 8,000 	& 50	&8	&5	\\
		Computer 	& 8,092	&100	&12 &5\\
		CalHouse	& 20,490 	& 150 &8&5\\ 
		Census 	& 22,584 	& 200 &16&5	\\ 	
		\bottomrule
	\end{tabular}
\end{table}
Each benchmark dataset is randomly split into $20$ partitions.

\subsubsection{Metrics}
We use \ac{MAE} to measure the efficacy: 
\begin{equation*}
MAE = \frac{1}{N} \sum_{n=1}^{N} |y_n-\hat{y}_n|,
\end{equation*}
where $\hat{y}_n$ is the predicted category.
As for efficiency, we choose running time (in seconds) as the measurement. 

\subsubsection{Methods used for comparison}
We choose \ac{KDOR}, \ac{GPOR}, \ac{SVOR}, \ac{SBOR} and \ac{ISBOR} discussed in the related work section as baselines. 
We use the ORCA package~\cite{ORsurvey} (in MATLAB)\footnote{\url{https://github.com/ayrna/orca}} for \ac{KDOR}. 
The authors of \ac{SVOR} and \ac{GPOR} provide a publicly available implementation in C.\footnote{\url{http://www.gatsby.ucl.ac.uk/~chuwei/\#software}} 
We use a MATLAB implementation of \ac{ISVOR} shared by the authors. 
\ac{SBOR} and \ac{ISBOR} are implemented in MATLAB.

\subsubsection{Settings and parameters}
We choose the Gaussian RBF in Eq.~\eqref{eq: kernel} as the basis function for each algorithm. 
We initialize \ac{ISBOR} by setting $\boldsymbol{\alpha} = 10^{-3}$, $\sigma=1$.\footnote{This is a heuristic setup inspired by \citet{GPOR}, although the better way to choose the starting points is by trying different values and selecting the best combination.}
We select the kernel width via $5$-fold cross-validation on the training set within the values of $\theta \in \{10^{-2}, 10^{-1}, \ldots, 10\}$. 
GPOR automatically learns the hyper-parameters, which does not require any pre-selection process. 
For other methods, we follow the model selection process in \cite{ORsurvey} and use a nested $5$-fold cross-validation on the training set to search for the best hyper-parameters. 
Specifically, we choose $\theta \in \{10^{-3}, 10^{-2}, \ldots, 10^3\}$ for every algorithm.
The additional regularization parameter of SVOR and ISVOR are chosen within the values of $c \in \{10^{-1}, \dots,10^3\}$. 
For \ac{KDOR}, we choose the regularization parameter within the range of $c \in\{0.1, 1, 10\}$, since the regularization parameter of KDOR presents a different interpretation from the one in \ac{SVM}. 
Additionally, \ac{KDOR} requires another singularity-avoiding parameter, which is chosen in the range of $u \in \{10^{-6}, 10^{-5}, \ldots, 10^{-1}\}$. 

Cross-validation is conducted using \ac{MAE}. 
That is, once the hyper-parameters with the lowest \ac{MAE} are obtained, we apply them to the whole training set and then validate them on the test sets. 

The experiments are run on a server with Intel(R) Xeon(R) CPU E5-2683 v3 2.00GHz (16 Cores) and 32 Gigabyte. 

\subsection{Experimental results}

\subsubsection{Efficacy}
\begin{figure}[h]
	\centering
	\subfigure[\ac{MAE}.]{
		\includegraphics[clip, trim=0mm 0mm 0mm 10mm,width = 0.48\textwidth]{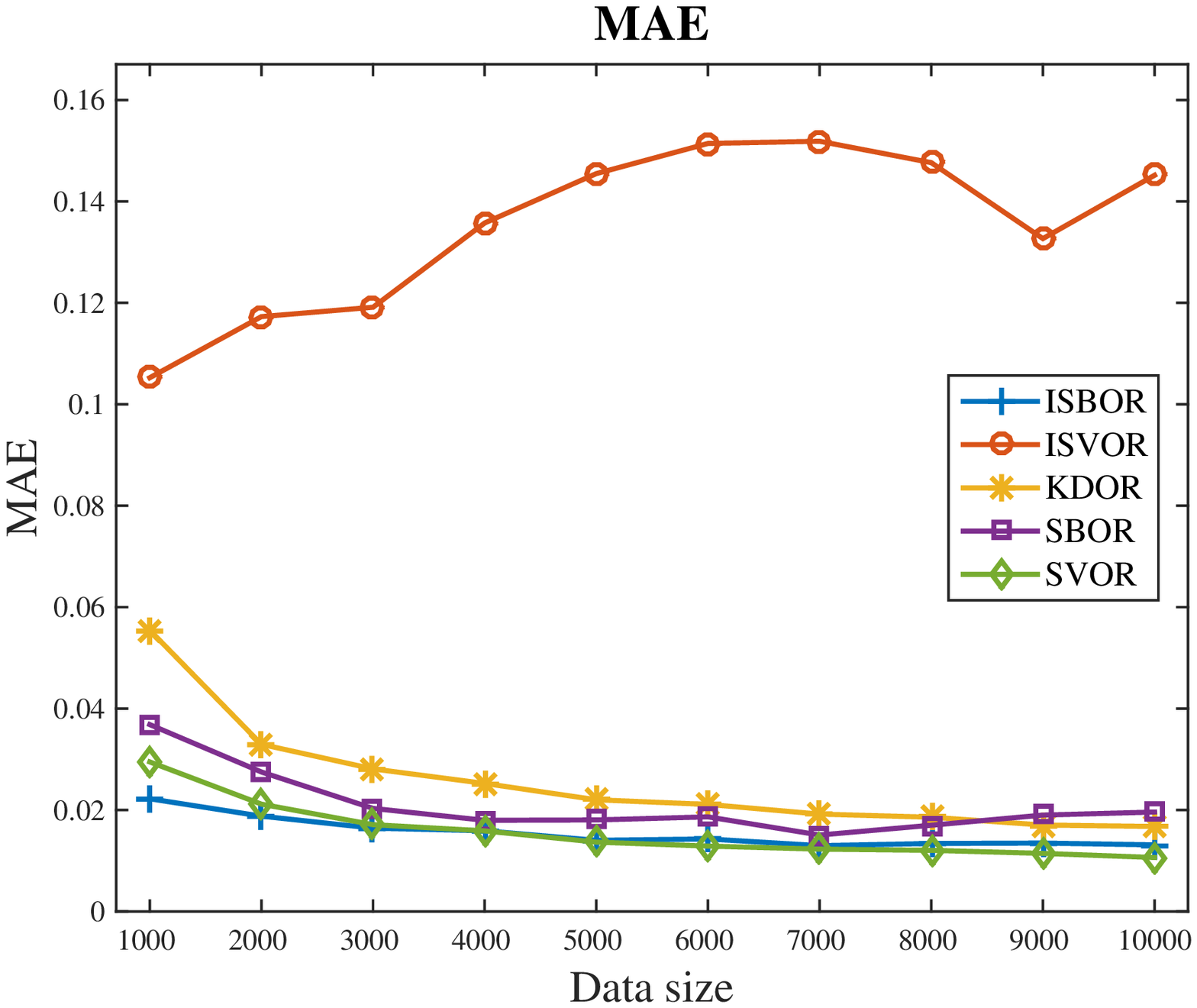}
		\label{fig:MAE}
		}
	\subfigure[Running time.]{
		\includegraphics[clip, trim=0mm 0mm 0mm 10mm,width = 0.48\textwidth]{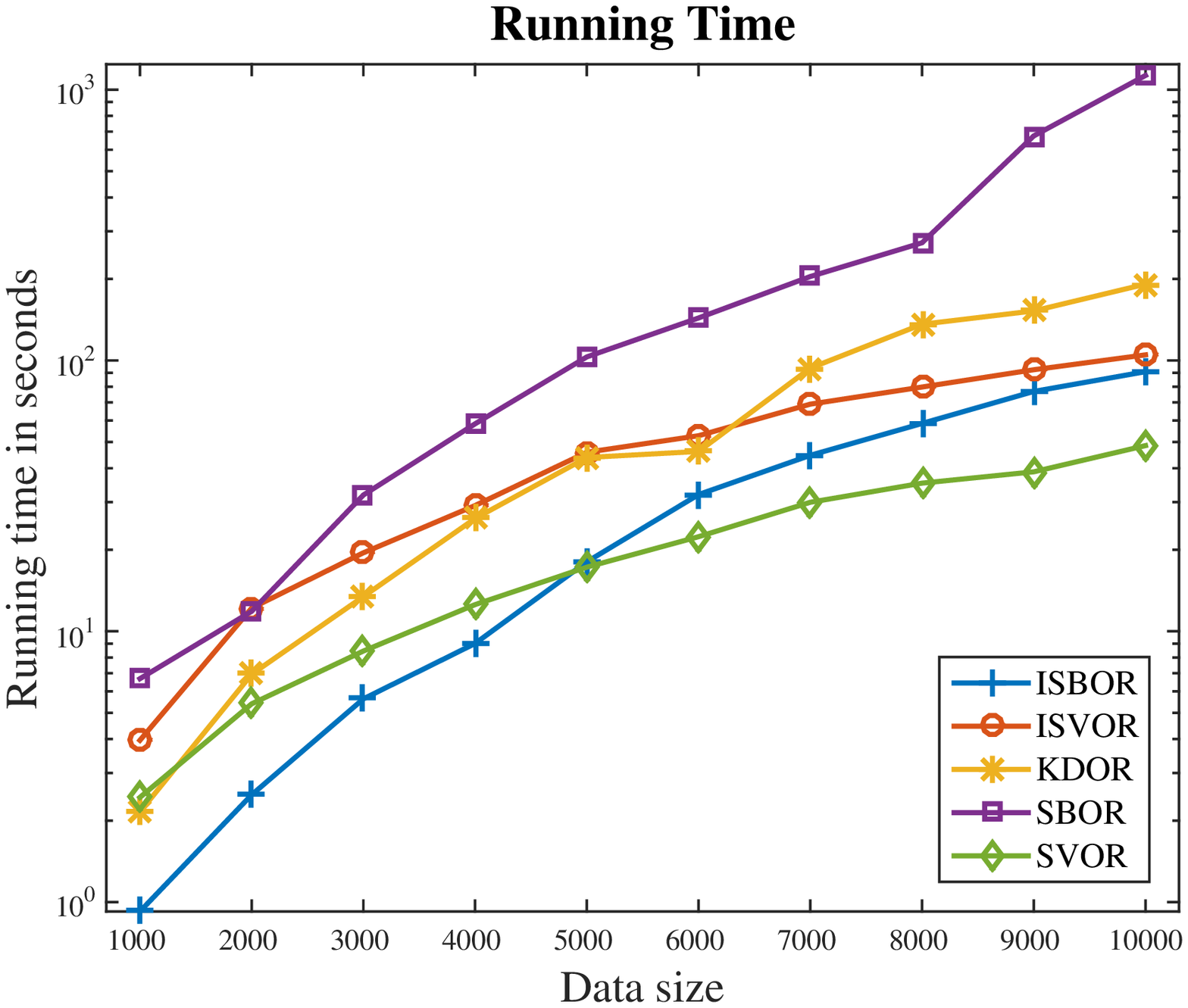}
		\label{fig:Runningtime}
		}
	\caption{\ac{MAE} and running time of \ac{OR} algorithms on the synthetic dataset.}
	\label{fig:synthetic}
\end{figure}

We begin by addressing \textbf{RQ1} concerning efficacy.
We first consider the results on the synthetic dataset.
Figure~\ref{fig:MAE} shows the performance in terms of \ac{MAE} on the synthetic dataset.
From the figure, we see that other than \ac{ISVOR}, all the algorithms work well on the Synthetic datasets, in terms of efficacy. 
Specifically, \ac{ISBOR} and \ac{SVOR} are the two best performing algorithms. 
When the data sizes are larger than $5000$, \ac{SVOR} outperforms \ac{ISBOR}, but the gaps are small. 

Next, we turn to the benchmark datasets.
The \ac{MAE} scores are presented in Table~\ref{tb:benchmarksResults} (top half). The results are averaged over $20$ partitions. 

\begin{table*}
	\centering
	\caption{Benchmark results: MAE and running time. Standard deviations (of MAE) indicated in brackets. Failure to complete all runs in $24$ hours is indicated with `--'; best results are marked in \textbf{boldface}, second best in \emph{italics}.}
	\label{tb:benchmarksResults}
	\small
	\begin{tabular}{l c c c c c c c }
		\toprule
		\emph{MAE}	&BS &SWD &Market & Bank & Computer & CalHouse & Census \\
		\midrule
		KDOR & 0.17 (0.03) & 0.58 (0.03) & 1.60 (0.03)& 0.21 (0.07) & 0.39 (0.03) &\textbf{0.46 (0.05)} & 0.63 (0.06) \\
		GPOR & 0.03 (0.02) & \emph{0.41 (0.03)} & -- & -- & -- & -- & -- \\
		SVOR &\textbf{0.00 (0.00)} 	&\textbf{0.41 (0.03)} &\textbf{0.83 (0.01)} & \emph{0.20 (0.06)} & 0.40 (0.03) &0.63 (0.05) & -- \\
		SBOR & 0.04 (0.06) & 0.52 (0.06) & {1.45 (0.03)} & 0.34 (0.19) & 0.44 (0.12) & -- &\textbf{0.60 (0.30)}\\
		ISVOR &0.36 (0.53) & 0.56 (0.04)&\emph{1.20 (0.08)} & 0.80 (0.10) &\textbf{0.36 (0.04)}&0.87 (0.07) &0.65 (0.06)\\
		ISBOR & \emph{0.02 (0.01)} & {0.43 (0.02)} & 1.74 (0.05) & \textbf{0.19 (0.05)} & \emph{0.37 (0.05)} & \emph{0.50 (0.04)} & \emph{0.63 (0.05)} \\			
		\midrule
		\emph{Running} \smash{\rlap{\emph{time}}} &BS &SWD &Market & Bank & Computer & CalHouse & Census \\
		\midrule
		KDOR	&\textbf{0.08}	&\textbf{0.27}	&\smash{\llap{1}}59.92		&97.27	&96.86		&\smash{\llap{1,}}369.14	&\smash{\llap{1,}}696.18	\\
		GPOR	&\smash{\llap{35}}9.94	&\smash{\llap{20}}5.60 		&--	& -- & -- & -- & --\\
		SVOR 	&\textbf{0.08}	&0.83 	&44.49	&\smash{\llap{9}}32.59 	& \smash{\llap{2,6}}82.19 &\smash{\llap{4,3}}50.30 & -- \\
		SBOR 	&0.96		&2.96	& 93.73		& \smash{\llap{9}}86.12		&\smash{\llap{2}}04.85	& -- & \smash{\llap{3,}}713.62	\\
		ISVOR 	&1.53	&\emph{0.77}	&\emph{65.22} 	&\textbf{73.89} &\textbf{73.61} &\emph{907.95}&\emph{774.22}\\			
		ISBOR	&0.64 &1.35	&\textbf{62.76}	&\emph{91.02}		&\emph{94.84}	&\textbf{810.22}		&\textbf{710.84}\\
		\bottomrule
	\end{tabular}
\end{table*}

To determine the significance of observed differences, we use the Wilcoxon test~\cite{wilcoxon1945individual,demvsar2006statistical} and compare the efficacy of each pair of algorithms. 
Since we compare $6$ algorithms, there are $30$ comparisons for each dataset in total. 
We choose the significance level $\alpha = 0.1$ and take the number of comparisons into account, and obtain the corrected significance level as $\alpha=0.1/30 \approx 0.0033$. 
For each algorithm, we record the number of statistically significant wins, losses (or failures in finishing the training on time) and draws. 
The Wilcoxon test results are reported in Table~\ref{tb:Wilcoxon}. 

\begin{table}[h]
	\caption{Wilcoxon tests for the \ac{MAE} results obtained using the benchmark datasets and reported in Table~\ref{tb:benchmarksResults}.}
	\label{tb:Wilcoxon}
	\centering
	\begin{tabular}{l@{~~} r r r} 
		\toprule
		Method 	& \# wins 	& \# draws & \# losses \\ 
		\midrule
		\ac{GPOR} & 4 & 6 & 25	\\
		\ac{SVOR} & 24 & 7 & 4 \\
		\ac{SBOR} &11 & 12 & 12\\
		\ac{KDOR} &11 & 8 & 16 \\
		\ac{ISVOR} &11 & 11 & 13 \\
		\ac{ISBOR} &17 & 10 & 8 \\
		\bottomrule
	\end{tabular}
\end{table}

Based on the top half of Table~\ref{tb:benchmarksResults} and Table~\ref{tb:Wilcoxon}, we find that \ac{SVOR} is the best performing ordinal regression algorithm in terms of \ac{MAE}. 
Specifically, \ac{SVOR} wins $24$ times out of $35$ pair-wise comparisons. 
\ac{ISBOR}, the second best performing algorithm, wins $17$ comparisons. 
Because of the time limitation, \ac{GPOR} fails to complete the experiments on $5$ datasets and performs worse. 
The rest algorithms performs similar with each others and win $11$ times. 

To sum up, these results answer \textbf{RQ1} as follows: although \ac{SVOR} has the best generalization performance, \ac{ISBOR} outperforms other baselines and is comparable to \ac{SVOR}.

\subsubsection{Efficiency}
We turn to \textbf{RQ2}. We report the running time of competing algorithms on the synthetic dataset with different data scales in Figure~\ref{fig:Runningtime}. 
Generally, the implementations in C run much faster than those in pure MATLAB. 
To suppress this effect, we compare the running times on a logarithmic scale. 
We omit plotting the results of \ac{GPOR}, because after running $24$ hours \ac{GPOR} failed to complete any run on any partition. 

Considering Figure~\ref{fig:Runningtime}, when it comes to efficiency, \ac{ISBOR} is faster than all algorithms except for \ac{SVOR}, which is implemented in C. 
Comparing to \ac{SBOR}, which can be regarded as the offline version of \ac{ISBOR}, the gaps between \ac{ISBOR} and \ac{SBOR} are getting larger with the size of data increasing. 
On $10000$-size data, \ac{ISBOR} is about $10$ times faster than \ac{SBOR}. 
These results demonstrate that incremental learning together with the sparseness assumption can accelerate the training speed of \ac{ISBOR}. 
In summary, Figure~\ref{fig:synthetic} shows that \ac{ISBOR} can be an efficient ordinal regression algorithm while preserving a comparable prediction accuracy to \ac{SVOR}. 

From the bottom part of Table~\ref{tb:benchmarksResults}, we notice that on the small datasets, \ac{ISBOR} does not show any advantages in running time. 
However, on the large datasets, \ac{ISBOR} outperforms the baselines. 
Specifically, we can see a trend that the larger scale of the dataset is, the bigger the gaps between \ac{ISBOR} and the batch algorithms are. 
This trend provides an answer to \textbf{RQ2}: the incremental setting makes \ac{ISBOR} a faster \ac{OR} algorithm. 

\subsubsection{Sparseness}
Finally, we address \textbf{RQ3}.
Since \ac{GPOR} and \ac{KDOR} make predictions based on all training samples, in Table~\ref{tb:svrv}, we only report the number of support or relevant samples of \ac{SVOR}, \ac{ISVOR}, \ac{SBOR} and \ac{ISBOR} so as to answer the sparseness question (\textbf{RQ3}). 

\begin{table}[h]
	\caption{Relevant and support samples used on the Benchmark datasets. Best results marked in \textbf{boldface}, second best in \emph{italics}.}
	\label{tb:svrv}
	\centering
	\begin{tabular}{ l c c c c }
		\toprule
		Dataset	 	&\ac{SVOR} 			&\ac{ISVOR} 	&\ac{SBOR} &\ac{ISBOR}\\
		\midrule
		BS & \phantom{00,0}60.2 & \phantom{00,}283.3 & \textbf{\phantom{0,00}9.0} & \emph{17}\phantom{.0}\\
		SWD & \phantom{00,}718.9	& \phantom{00,}454.1 & \phantom{0,}\emph{104.8} & \textbf{58.5}\\
		Marketing & \phantom{0}3,756.0 & 10,185.1 & \phantom{0,0}\emph{51.3} & \textbf{51.1}\\
		Bank 		 & \phantom{0}5,685.1 & \phantom{0}8,128.5 & \phantom{0,0}\textbf{16.6} & \emph{44.8}\\
		Computer 		&\phantom{0}3,373.1 & \phantom{0}7,739.5 & \emph{3,056.1} & \textbf{30.9}\\
		CalHouse & 12,788.3 & \emph{23,919.8} & -- & \textbf{84.8}\\
		Census & --& 28,348.5 & \emph{1,001.0} & \textbf{73.0}\\
		\bottomrule
	\end{tabular}
\end{table}

\noindent%
Analyzing Table~\ref{tb:svrv}, we notice that the sparse Bayes based \ac{SBOR} and \ac{ISBOR} employ much smaller numbers of training samples to make predictions than the \ac{SVM}-based \ac{SVOR} and \ac{ISVOR}.\footnote{Notice how \ac{ISVOR} uses more samples than the ground truth provides due to binary decomposition, as explained in Section~\ref{sec:relateWork}.} 
Among the seven benchmark datasets, \ac{ISBOR} wins $5$ times and \ac{SBOR} wins $2$ times, which supports our claim that \ac{ISBOR} is a parsimonious ordinal regression algorithm and can make effective predictions based on a small subset of the training set. 
This finding answers \textbf{RQ3} on sparseness.

\section{Conclusion}
\label{sec:conclusion}
We have presented a novel incremental ordinal regression algorithm within an efficient sparse Bayesian learning framework. 
Instead of processing the whole training set in one go, the proposed algorithm can incrementally learn from representations of training samples and has linear computational complexity in the training data size. 
Our empirical results show that \acf{ISBOR} is comparable or superior to state-of-the-art \ac{OR} algorithms based on basis functions in terms of efficacy, efficiency and sparseness. 

We hope that this work paves the way for research into large-scale ordinal regression.  
We believe that the design of \ac{ISBOR} can be improved in multiple directions. 
From a Bayesian viewpoint, a more elegant way to optimize the hyper-parameters would be to maximize $p(\boldsymbol{\eta}\mid \mathbf{D})$ rather than $p(\mathbf{D} \mid \boldsymbol{\eta})$ with additional hyper-assumptions. 
This is achievable via other approximation inference methods like variational Bayes and expectation propagation~\cite[Chapter~10]{prml}. 
From an application view, we can equip \ac{ISBOR} with other sparse Bayesian architectures and adapt it to other problems like semi-supervised learning~\cite{semiORsvm,elsevier3,srijith2013semi} and feature selection~\cite{jiang2016probabilistic,FPCVM}. 
From a ranking viewpoint, higher positions are more important. 
So far, \ac{ISBOR} ignores pair-wise preferences and considers each position equally important, which amounts to a point-wise approach. 
Another promising future direction, therefore, is to take pair-wise position information into account and apply \ac{ISBOR} to ranking problems.

\subsection*{Code and data}
\label{sec:code}
To facilitate reproducibility of the results in this paper, we are sharing the code and the data used to run the experiments in this paper at \url{https://github.com/chang-li/SBOR}.

\subsection*{Acknowledgments}
We thank our anonymous reviewers for their valuable feedback and suggestions.

This research was partially supported by
Ahold Delhaize,
Amsterdam Data Science,
the Bloomberg Research Grant program,
the China Scholarship Council,
the Criteo Faculty Research Award program,
Elsevier,
the European Community's Seventh Framework Programme (FP7/2007-2013) under
grant agreement nr 312827 (VOX-Pol),
the Google Faculty Research Awards program,
the Microsoft Research Ph.D.\ program,
the Netherlands Institute for Sound and Vision,
the Netherlands Organisation for Scientific Research (NWO)
under pro\-ject nrs
CI-14-25, 
652.\-002.\-001, 
612.\-001.\-551, 
652.\-001.\-003, 
and
Yandex.
All content represents the opinion of the authors, which is not necessarily shared or endorsed by their respective employers and/or sponsors.
	
\section*{References}
\bibliographystyle{elsarticle-num-names}
\bibliography{sbor-chang}

\end{document}